**Deep learning models for predicting RNA degradation via dual crowdsourcing**

Hannah K. Wayment-Steele*[1,5], Wipapat Kladwang*[2,5], Andrew M. Watkins*[2,5], Do Soon Kim*[2,5], Bojan Tunguz*[2,3], Walter Reade[4], Maggie Demkin[4], Jonathan Romano[2,5,6], Roger Wellington-Oguri[5], John J. Nicol[5], Jiayang Gao[7], Kazuki Onodera[8], Kazuki Fujikawa[9], Hanfei Mao[10], Gilles Vandewiele[11], Michele Tinti[12], Bram Steenwinckel[11], Takuya Ito[13], Taiga Noumi[14], Shujun He[15], Keiichiro Ishi[16], Youhan Lee[17], Fatih Öztürk[18], Anthony Chiu[19], Emin Öztürk[20], Karim Amer[21], Mohamed Fares[22], Eterna Participants**[5], Rhiju Das[2,5,23]

[1]Department of Chemistry, Stanford University, Stanford, California 94305, USA
[2]Department of Biochemistry, Stanford University, California 94305, USA
[3]NVIDIA Corporation, Santa Clara, California 95051
[4]Kaggle, San Francisco, California 94107
[5]Eterna Massive Open Laboratory
[6]Department of Computer Science and Engineering, State University of New York at Buffalo, Buffalo, New York, 14260, USA
[7]High-flyer AI, Hangzhou, Zhejiang, China, 310000
[8]NVIDIA Corporation, Minato-ku, Tokyo 107-0052, Japan
[9]DeNA, Shibuya-ku, Tokyo 150-6140, Japan
[10]Yanfu Investments, Shanghai, China, 200000
[11]IDLab, Ghent University, Technologiepark-Zwijnaarde, Gent, Belgium, B-9052
[12]College of Life Sciences, University of Dundee, Dundee DD1 4HN, United Kingdom
[13]Universal Knowledge Inc., Tokyo 150-0013, Japan
[14]Keyence Corporation, 1-3-14, Higashi-Nakajima, Higashi-Yodogawa-ku, Osaka, 533-8555, Japan
[15]Department of Chemical Engineering, Texas A&M University, College Station, TX 77843
[16]Rist Inc, Meguro-ku, Tokyo 153-0063, Japan
[17]Kakao Brain, Seongnam, Gyeonggi-do, Republic of Korea
[18]H2O, Istanbul, 3400, Turkey
[19]Clover Health, Hong Kong, 999077, PRC
[20]Afiniti, Istanbul, 3400, Turkey
[21]Center for Informatics Science, Nile University, Sheikh Zayed, Giza, Egypt, 12588
[22]National Research Centre, Dokki, Cairo, Egypt, 12622
[23]Department of Physics, Stanford University, California 94305, USA

* these authors contributed equally.
** Eterna consortium authors are listed in Table S1.

## Abstract

Messenger RNA-based medicines hold immense potential, as evidenced by their rapid deployment as COVID-19 vaccines. However, worldwide distribution of mRNA molecules has been limited by their thermostability, which is fundamentally limited by the intrinsic instability of RNA molecules to a chemical degradation reaction called in-line hydrolysis. Predicting the degradation of an RNA molecule is a key task in designing more stable RNA-based therapeutics. Here, we describe a crowdsourced machine learning competition ("Stanford OpenVaccine") on Kaggle, involving single-nucleotide resolution measurements on 6043 102-130-nucleotide diverse RNA constructs that were themselves solicited through crowdsourcing on the RNA design platform Eterna. The entire experiment was completed in less than 6 months, and 41% of nucleotide-level predictions from the winning model were within experimental error of the ground truth measurement. Furthermore, these models generalized to blindly predicting orthogonal degradation data on much longer mRNA molecules (504-1588 nucleotides) with improved accuracy compared to previously published models. Top teams integrated natural language processing architectures and data augmentation techniques with predictions from previous dynamic programming models for RNA secondary structure. These results indicate that such models are capable of representing in-line hydrolysis with excellent accuracy, supporting their use for designing stabilized messenger RNAs. The integration of two crowdsourcing platforms, one for data set creation and another for machine learning, may be fruitful for other urgent problems that demand scientific discovery on rapid timescales.

**Introduction**

Therapeutics based on messenger RNA (mRNA) have shown immense promise as a modular therapeutic platform, allowing potentially any protein to be delivered and translated[1, 2], as evidenced by the rapid deployment of mRNA-based vaccines against SARS-CoV-2[3-5]. However, the chemical instability of RNA sets a fundamental limit on the stability of RNA-based therapeutics such as mRNA-based vaccines[1, 6-8]. RNA hydrolysis is the limiting factor in lipid nanoparticle (LNP)-based formulations[9, 10]. Hydrolysis in LNP formulations degrades the amount of mRNA remaining during shipping and storage, and hydrolysis in vivo after vaccine injection limits the amount of resulting protein produced over time[9]. Better methods to develop thermostable RNA therapeutics would allow for increasing the equitability of their distribution, reducing their cost, and possibly increasing their potency[10, 11]. Initial results have demonstrated that more stable mRNAs can be designed via stochastic algorithms that identify mRNA sequences that code for the same protein[12], but are predicted to be more resistant to hydrolysis[13, 14]. Importantly, these mRNAs are demonstrated to produce equivalent amounts of protein to non-optimized mRNAs[14]. These design strategies are predicted to be able to produce mRNAs that do not activate double-stranded RNA immune sensors[13, 15], and have also demonstrated to be compatible with mRNAs synthesized from modified nucleotides including pseudouridine[14], which are used in mRNA vaccine formulations[16].

A key prediction task underpinning algorithms for designing mRNAs with increased stability is the model used to predict of RNA hydrolysis from sequence (Figure 1A). Previous models for RNA degradation have assumed that the probability of any RNA nucleotide linkage being cleaved is proportional to the probability of the 5' nucleotide being unpaired[13]. Computational studies with this model suggested that at least a two-fold increase in stability could be achieved through sequence design, while maintaining a wide diversity of sequences and features related to translatability, immunogenicity, and global structure[14]. However, it is unlikely that degradation depends only on the probability of a nucleotide being unpaired: local sequence-and structure-specific contexts may vary widely, as evidenced by ribozyme RNAs found in nature, whose sequences adopt specific structures that undergo self-scission[17].

We wished to understand the maximum predictive power achievable for RNA degradation on a short timescale for model development. To do this, we combined two crowdsourcing platforms: Eterna, an RNA design platform, and Kaggle, a platform for machine learning competitions. Eterna has previously been able to solve near-intractable problems in RNA design[18, 19], and the diversity of resulting structures on its platform have more recently contributed to advancing RNA secondary structure prediction[20]. We reasoned that crowdsourcing the problem of obtaining data on a wide diversity of sequences and structures would rapidly lead to a diverse dataset, and that crowdsourcing the second problem of obtaining a machine learning architecture would result in a model capable of expressing the resulting complexity of sequence- and structure-dependent degradation patterns. We hypothesized this "dual crowdsourcing" would lead to stringent and independent tests of the models developed, minimizing interplay between the individuals designing the constructs to test (Eterna participants) and the individuals building the models (Kaggle participants) and leading to better generalizability on independent data sets.

The resulting models were subjected to two blind prediction challenges. The first was in the context of the Kaggle competition, where the RNA structure probing and degradation data that participants would be aiming to predict was not acquired until after the competition was announced. The experimental method used for these data, In-line-seq, allowed for measuring the degradation rate of individual dinucleotide linkages. However, this method relies on probing short RNA fragments and is unable to scale to make single-nucleotide degradation measurements of full-length mRNAs for protein targets of interest. Other experimental methods such as PERSIST-seq[14] have been developed to characterize the overall degradation rates per mRNA molecule, which is the primary value of interest to minimize when designing stabilized RNA-based therapeutics. In principle, the overall degradation rate of a mRNA molecule of length $N$ is equivalent to the sum of degradation rates at each dinucleotide linkage in the backbone[21]:

$$k_{deg}^{mRNA} = \sum_{i=1}^{N-1} k_{deg}^i, \quad (1)$$

Where $k_{deg}^i$ is the degradation of nucleotide linkage $i$. The half-life of the mRNA is calculated as

$$t_{1/2} = \frac{\ln 2}{k_{deg}^{mRNA}} \quad (2).$$

We tested the above model empirically by comparing the summed degradation rates per nucleotide to the log abundance of the entire construct remaining from sequencing and found high agreement (Figure S1). Using the above ansatz, the resulting models were tested in a second blind challenge of predicting the overall degradation of full-length mRNAs encoding a variety of model proteins, experimentally tested using PERSIST-seq. The models also demonstrated increased predictive power over existing methods in predicting these overall degradation rates. These models therefore appear immediately useful for guiding design of low degradation mRNA molecules. Analysis of model performance suggests that the task of predicting RNA degradation patterns is limited by both the amount of data available as well as the accuracy of the structure prediction tools used to create input features. Further developments in experimental data and secondary structure prediction, when combined with network architectures such as those developed here, will further advance RNA degradation prediction and therapeutic design.

**Results**

*Dual-crowdsourced competition design and assessment.* The aim of the OpenVaccine Kaggle competition (Figure 1B) was to develop computational models for predicting RNA degradation patterns. We asked participants on the Eterna platform to submit RNA designs using a web-browser design window (Figure 1C), which resulted in a diversity of sequences and structures (Figure 1D). 150 participants in total (Table S1) submitted sequences. A secondary motivation was an opportunity for participants to receive feedback on RNA fragments they may wish to use in mRNA design challenges described in Leppek et al.[14] 3029 RNA designs of length 107 nt were collected in the first "Roll-Your-Own-Structure" round I (RYOS-I), which was opened March 26, 2020, and closed upon reaching 3029 constructs on June 19, 2020 (Figure 1E).

We then obtained nucleotide-level degradation profiles for the first 68 nucleotides of these RNAs using In-line-seq[14], a novel method for characterizing in-line RNA degradation in high-throughput for the purposes of designing stabilized RNA therapeutics. Degradation profiles were collected in four different accelerated degradation conditions, and the structures of the constructs were also characterized via selective 2' hydroxyl acylation with primer extension (SHAPE; termed "Reactivity" below)[22, 23], a technique to characterize RNA secondary structure. The Kaggle competition was designed to create models that would have predictive power for three of these data types, given RNA sequence and secondary structure as input (Figure 1F). Though SHAPE reactivity is distinct from the degradation measurements we wished to be able to predict for the application of mRNA design, the SHAPE reactivity datasets had higher signal-to-noise ratio (Figure S2), and we hypothesized that models would demonstrate improved performance predicting SHAPE reactivity and that the data would serve as a "positive control" for a predictable data type. Additionally, models with highly accurate predictions of SHAPE reactivity would have utility in biological applications.[24]

In total, each independent construct of length N required predicting 3xN values for the 3 data types. In addition to these experimental data, Kaggle participants were also provided with features related to RNA secondary structure computed from available biophysical models to use if they wished. These features included 107x107 base pairing probability matrices from EternaFold[20], a recently developed package with state-of-the-art performance on RNA structural ensembles; dot-parenthesis notated minimum free energy (MFE) RNA secondary structure from the ViennaRNA package[25]; and a six-character featurization of the MFE structure calculated using bpRNA[26].

We developed training and "public test" datasets from the RYOS-I dataset (Figure 2). The public test dataset was used to rank submissions during the competition. The 3029 constructs were filtered for those with mean signal-to-noise values greater than 1, resulting in 2218 constructs (Figure 2, dark blue track, see Methods). These constructs were segmented into splits of 1179 in the public training dataset, 400 constructs in the public test set, and 639 for the "private test" dataset, the set which would be used in the final evaluation. The sequences that did not pass the signal-to-noise filter were also provided to Kaggle participants with the according description. The RYOS-I data contained some "clusters" of sequences where Eterna players included many small variations on a single sequence (clusters visible in Figure 1D). To mitigate the possibility of sequence motifs in these clusters biasing evaluation, we segmented the RYOS-I data into a public training, public test, and private test sets by clustering the sequences and including only sequences that were singly, doubly, or triply-clustered in the private test set (see Methods). This strategy was described to Kaggle participants during the competition.

To ensure that the majority of the data used for the private test set was fully blind, we initiated a second "Roll-your-own-structure" challenge that was launched for Eterna design collection on August 18, 2020. Given that useful models for degradation should be agnostic to RNA length, we designed the constructs in RYOS-II to be 34 nucleotides longer (102 vs. 68 nts) than the constructs in RYOS-I to discourage modelling methods that would overfit to constructs of length 68. Design collection was closed on September 7th, three days before the launch of the Kaggle challenge on September 10th. The RYOS-II wet-lab experiments were conducted concurrently with the Kaggle challenge, enabling a completely blind test for the models developed on Kaggle. The Kaggle competition was closed on October 6th. The RYOS-II was similarly clustered and filtered to ensure that the test set used for scoring consisted primarily of

singly- and double-clustered constructs. Three data types were used to score models: SHAPE; 10 mM Mg$^{2+}$, pH 10, 1 day, 24 °C; and 10 mM Mg$^{2+}$, pH 7.2, 1 day, 50 °C. Models were scored using the mean column RMSE (MCRMSE) across three data types, defined as

$$MCRMSE = \frac{1}{N_t}\sum_{j=1}^{N_t}\sqrt{\frac{1}{n}\sum_{i=1}^{n}(y_{ij} - \hat{y}_{ij})^2}, \qquad (3)$$

Where $N_t$ is the number of scored data types, $n$ is the number of nucleotides in the dataset, and t, $y_{ij}$ is the measured data value, and $\hat{y}_{ij}$ is the predicted data value. Two additional data types were included in the training data corresponding to RNAs degraded for 7 days rather than 1 day without Mg2+: (pH 10, 7 days, 24 °C; and pH 7.2, 7 days, 50 °C). However, these data were not collected for the second round to accelerate competition turnaround.

*Performance of Kaggle teams and common attributes of top-performing models.* During the three week competition period, 1,636 teams submitted 35,806 solutions. Overall performance of teams vs. baseline models for RNA degradation are depicted in Figure 3A. Kaggle entries significantly outperformed the "DegScore" linear regression model for RNA degradation[14] by 37 % in MCRMSE for the public test set and 25% for the private test set (Figure 3A). An additional benchmark model which used the DegScore windowed featurization (see Methods) with improved XGBoost[27] training, termed the DegScore-XGB model, resulted in moderate improvement (Public MCRMSE 0.35854, Private MCRMSE 0.43850). Kaggle participants developed feature encodings beyond what was provided. One of the most widely-used community-developed featurizations was a graph-based distance embedding depicted in Figure 3B. Several teams, including the top three teams, used a publicly-shared autoencoder/GNN/GRU kernel (https://www.kaggle.com/code/mrkmakr/covid-ae-pretrain-gnn-attn-cnn/), which alone achieved a MCRMSE of 0.24860 on the Public and 0.36106 on the Private test set (Figure 3A). This notebook was the most forked (forked 936 times as of March 2022) and upvoted (upvoted 386 times). The architecture of the winning "Nullrecurrent" model (Figure 3C) depicts the architecture of this shared kernel, which feeds 1D and 2D features, including adjacency matrices based on the secondary structure of the RNA inputs, into multi-head attention network that is then fed into convolutional neural net layers.. Many teams additionally cited pseudo-labeling and generating additional mock data as being integral to their solutions. The architecture of the second-place team (Figure 3D) demonstrates an example implementation of using pseudo-labeling. The machine learning practice of pseudo-labeling involves using predictions from one model as "mock ground truth" labels for another model. Effective pseudo-labeling usually requires a high level of accuracy of the primary model and is most frequently used with classification problems. To generate additional mock data, participants generated random RNAs as well as structure featurizations using 5 different secondary structure prediction algorithms using the package Arnie (https://github.com/DasLab/arnie) and iteratively scored based on these predictions for their model as well (see Supplement for more detailed descriptions of solutions from Kaggle teams).

*Ensembling models.* We explored whether increased accuracy in modeling might be achieved by combining models, as a common feature of Kaggle competitions is that winning solutions are dissimilar enough that ensembled models frequently improve predictive ability. We used a genetic algorithm to ensemble maximally 10 of the top 100 models. The score on the public dataset was used to optimize, with the final ensembled model evaluated on the private dataset. With this method, ensembling achieved a Public MCRMSE of 0.2237 (compared to the best Public MCRMSE of 0.2276) and a Private MCRMSE of 0.3397 (compared to the best Private test set MCRMSE of 0.3420). In comparison, averaging the outputs of the top two models gave a result of 0.2244 Public, 0.33788 Private. Blending the top two solutions with the 3rd solution did not improve the result. An estimated bound of ensembling can be found by optimizing directly to the Private ensemble score. With this method, it was possible to achieve a Private ensemble score of 0.3382 (again, vs. best Leaderboard MCRMSE 0.3420). The improvement of 0.0038 over the leaderboard for this last approach is about the distance between the 1st place and 10th place teams, and the "correct" way gives an improvement that is the distance between the 1st and 5th place teams. All these experiments suggest that most of the signal has been captured by the top two models, and that the use of further ensembling provides, at best, modest improvements. The seemingly puzzling result that the simple ensemble of the top two models outperforms the genetic algorithm blend of the top 10 (on the private test set) suggests that the genetic algorithm did not find a global minimum for model weights.

*Top models are capable of deep representation of RNA experimental motifs.* We analyzed predictions from the first-place model ("Nullrecurrent") in depth to better understand its performance. Across all nucleotides in the private test set, 41% of nucleotide-level predictions for SHAPE reactivity agreed with experimental measurements within an error that was lower than experimental uncertainty; for comparison, if experimental errors are distributed as normal distributions, a perfect predictor would agree with experimental values over 68% of data points. For Deg_Mg_pH10 and Deg_Mg_50C, 28% and 42% of predictions were within error, respectively. Per-construct RMSE between data types was moderately correlated, with the highest correlation being Spearman R=0.63 between deg_Mg_pH10 and deg_Mg_50C (Figure S3). The nucleotides with the highest RMSE in the Deg_Mg_pH10 data type were any nucleotide type in bulges, and U's in any unpaired context. Figure 4A depicts representative constructs with the lowest RMSE for the Deg_Mg_pH10 data type out of the private test data, demonstrating that a diverse set of structures and structure motifs were capable of being predicted correctly. Aggregating the predictions from the Nullrecurrent model over secondary structure motifs (Figure 4B) demonstrates that the Nullrecurrent predictions by motif captured patterns previously observed in the experimental signal[14]. The most reactive RNA structure motifs were triloops, a previously unknown biological finding. Another unexpected finding from these data were that internal loops with symmetric lengths on either side recovered stability over internal loops with asymmetric lengths. The fact that the Nullrecurrent model was able to capture this trend indicates that using such models within a design algorithm would allow for an automated way to impart such biological attributes within a designed mRNA. Constructs with the highest RMSE demonstrate indicators that the provided structure features were incorrect. Figure 4C depicts two constructs with the highest RMSE for the SHAPE modification prediction. The SHAPE data for the first construct, "2204Sept042020", has high reactivity in predicted stem areas, indicating the stems were unfolded in solution. In contrast, construct "Triple UUUU

Tetraloops" has experimentally low reactivity in the exterior loop, suggesting that a stem was present. However, we found no correlation between the EternaScore, a metric indicating how closely the experimental reactivity signal matches the predicted structure[18], and RMSE summed per construct for the private test constructs, suggesting that in general, quality of the input structure features was not a limitation in model training (Figure S4).

*Kaggle models show improved performance in independent mRNA degradation prediction.* As an independent test, we assessed the ability of the top two Kaggle models to predict the overall degradation rates of a dataset of full-length mRNAs, which were not publicly available at the time of the Kaggle competition. Because the throughput of the In-line-seq experimental method is limited to RNA lengths easily accessible by Illumina sequencers (500 nts), these mRNAs could not be probed at a per-nucleotide level akin to the datasets used in the Kaggle experiments. However, their overall degradation rates (related to the per-nucleotide degradation rate via Eqn. 1), were characterized using PERSIST-seq[14]. The lengths of these mRNAs ranged from 504 to 1588 with a median length of 928 (Figure 5A), nearly 10-fold times longer than the longest RNA fragments used in the OpenVaccine Kaggle challenge (full dataset, attributes, and calculations in Table S2). The experimentally-determined structures of two example mRNAs designed by Eterna participants,[14] which both code for Nanoluciferase but have a 2.5-fold difference in hydrolysis lifetime, are depicted in Figure 5B.

To compare the Kaggle predictors to the single overall degradation rate from PERSIST-seq, we made predictions for all nucleotides in the full mRNA constructs and summed the predictions from the region that was captured in the PERSIST-seq method by reverse-transcription PCR which, in most cases, included the mRNA's 5' untranslated region (UTR) and coding sequence (CDS) (Figure 5C). Carrying out predictions on the full RNA sequence and then summing over the probed window allows to account for interactions between the untranslated regions and CDS, as can be seen for two example constructs in Figure 5B -- nucleotides in the 5' and 3' UTRs are predicted to pair with the CDS. We made predictions for 188 mRNAs in 4 classes: a short multi-epitope vaccine (MEV), the model protein Nanoluciferase, with one class consisting of varied UTRs and a second consisting of varied CDSs, and enhanced Green Fluorescent Protein (eGFP). We found that the Kaggle second-place "Kazuki2" model exhibited the highest correlation to fit degradation rates, followed by the Kaggle 1st-place "Nullrecurrent" model (Figure 5C), with Spearman correlation coefficients of 0.48 (p=3.3e-12) and 0.43 (p=9.5e-10), respectively. Both Kaggle models outperformed unpaired probability values from ViennaRNA RNAfold v. 2.4.14[25] (R=0.25, p=5.4e-4), the DegScore linear regression model (R=0.36, p=2.9e-7) and the DegScore-XGBoost model (R=0.42, p=1.8e-09). An ensemble of the Nullrecurrent and Kazuki2 models did not outperform the Kazuki2 model (R=0.47, p=1.4e-11), again suggesting that the models themselves had reached their predictive potential. In comparison, resampling the measured degradation rates from within experimental error and calculating the correlation to the mean degradation rate, as a measure of the upper limit of experimental noise, resulted in a Spearman correlation of 0.88 (Table 1).

**Discussion**

The OpenVaccine competition uniquely leveraged resources from two complementary crowd-sourcing platforms: Kaggle and Eterna. The participants in the Kaggle competition were tasked with predicting stability measurements of individual RNA nucleotides. The urgency of timely development of a stable COVID-19 mRNA vaccine necessitated that the competition be run on a relatively short timeframe of three weeks, as opposed to three months, which is more common with the Kaggle competitions.

The models presented here are immediately useful for mRNA design in that they could be called within a stochastic mRNA design algorithm[21] to minimize the predicted degradation. There is likely further opportunity to leverage advancements in natural language processing to use datasets such as the ones presented here to generate candidate mRNA designs using text generation approaches [28-30]. The degradation data used in this competition were from unmodified nucleotides, but mRNA vaccines are being formulated with modified nucleotides including pseudouridine or N-1-methyl-pseudouridine [16]. Modified nucleotides in general will have differing underlying thermodynamics[12], and there is a need to develop datasets and predictive models to predict structures and resulting stabilization of mRNAs formulated with modified nucleotides. Short of developing complete new thermodynamic parameters for modified nucleotides, it may be possible to develop principled heuristics to adapt models to mRNAs synthesized with modified nucleotides. For instance, Leppek et al. modified the DegScore model for pseudouridine by setting all contributions to degradation from uracil to zero to mimic the stabilization effect, and saw moderate improvement in correlation[14].

Kaggle competitions with relatively small datasets can be subject to serious overfitting to the public leaderboard, which often leads to a major "shake up" of the leaderboard when the results on the unseen test set are announced. In this competition the shakeup was minimal - most of the top teams were ranked close to the same position on the private leaderboard as they were on the public leaderboard. As the private leaderboard was determined on data that had not been collected at the time of the competition launch, this result suggests that the models that were developed are robust and generalizable. Furthermore, the models generalized to the task of predicting degradation for full-length mRNA molecules that were ten-fold longer than the constructs used for training. We speculate that the use of a separate, independently collected data set for the private leaderboard tests -- a true blind prediction challenge -- was important for ensuring generalizability. The winning solutions all used neural network architectures that are commonly used with modeling of 1D sequential data: recurrent NNs (LSTMs and GRUs) and 1D CNNs. The effectiveness of pseudo-labeling has two implications: more data will likely benefit any future modeling efforts, and the simple neural networks that were used have enough capacity to benefit from more data.

An under-investigated aspect of the models presented here is the effect of training on multiple data types. We speculate that because SHAPE reactivity has higher signal-noise than the degradation data types (cf. Figure S2), models with architectures that allowed for weight-sharing between data types benefitted from learning to predict SHAPE reactivity as well. Directly predicting RNA degradation without concurrently training on SHAPE data may result in worse model performance. Conversely, the model architectures presented here may also prove to have useful biological applications in predicting only SHAPE reactivity data. Future directions for model development includes training such models on larger chemical mapping datasets from more diverse experimental sources[20], and integrating into inference frameworks for RNA structure prediction[20, 31].

Finally, the models for predicting RNA hydrolysis developed in this work may prove useful in computationally identifying classes of natural RNAs which have evolved to be resistant to degradation. Such future bioinformatic analysis may suggest entirely new biologically-inspired approaches for designing hydrolysis-resistant RNA therapeutics.

**Acknowledgments.** We thank all participants of the Kaggle OpenVaccine challenge. We thank Sharif Ezzat and Camilla Kao for Eterna development and assistance launching the OpenVaccine challenge. We acknowledge funding from the National Institutes of Health (R35 GM122579 to R.D.), FastGrants, and gifts to the Eterna OpenVaccine project from donors listed in Table S3.

**Contributions.** HKWS, DSK, AMW, BT, WR, MT, and RD designed and implemented the Kaggle OpenVaccine competition. WK performed all In-line-seq experiments. HKWS, DSK, AMW, RD designed the curated datasets. HKWS performed model analysis. HKWS, AMW, BT, RD wrote the manuscript. JG, KO, KF, HM, GV, MT, BS, TI, TN, SH, KI, YL, FO, AC, EO, KA, MF contributed as members of gold-winning teams from the Kaggle competition and wrote supplemental solution descriptions.

**Conflict of Interest.** DSK, WK, and RD hold equity and are employees of a new venture seeking to stabilize mRNA molecules. WR and MD are employees of Kaggle.

**Methods**

*Initial feature generation.* As a starting point for Kaggle teams, we supplied a collection of features for each RNA sequence, including the minimum free energy (MFE) structure according to the ViennaRNA 2 energy model[25], loop type assignments generated with bpRNA[26] (S=Stem, E=External

Loop, I=Internal loop, B=Bulge, H=Hairpin, M=Multiloop, X=Dangle) and the base pair probability matrix according to the EternaFold[20] energy model. These features were generated using Arnie (https://github.com/DasLab/arnie).

*Experimental data generation.* The first experimental dataset used in this work, for the public training and test set, resulted from the "Roll-Your-Own-Structure" Round I lab on Eterna, and had been generated previously by Leppek et al[14].

The second experimental dataset used in this work, for the private test set, was generated for this work specifically. To produce these data, and for precise consistency with the public training and test set, In-line-seq was carried out as described by Leppek et al.[14] In brief, DNA templates were ordered via custom oligonucleotide pool from Custom Array/Genscript, prepended by the T7 RNA polymerase promoter. Templates were amplified via PCR, transcribed to RNA via the TranscriptAid T7 High Yield Transcription Kit (Thermofisher, K0441), and the purified RNA was subjected to degradation conditions: 1) 50 mM Na-CHES buffer (pH 10.0) at room temperature without added MgCl2; 2) 50 mM Na-CHES buffer (pH 10.0) at room temperature with 10 mM $MgCl_2$; 3) phosphate buffered saline (PBS, pH 7.2; Thermo Fisher Scientific-Gibco 20012027) at 50˚C without added $MgCl_2$; and 4) PBS (pH 7.2) at 50˚C with 10 mM $MgCl_2$. In parallel, purified RNA was subjected to SHAPE structure probing conditions, and one sample was subjected to the SHAPE protocol absent addition of the 1-methyl-7-nitroisatoic anhydride reagent.

cDNA was prepared from the six RNA samples (SHAPE probed, control reaction, and four degradation conditions). We pooled 1.5 μL of each cDNA sample together, ligated with an Illumina adapter, washed, and resuspended the ligated product, which was quantified by qPCR, sequenced using an Illumina Miseq. Resulting reads were analyzed using MAPseeker (https://ribokit.github.io/MAPseeker) following the recommended steps for sequence assignment, peak fitting, background subtraction of the no-modification control, correction for signal attenuation, and reactivity profile normalization as described in Seetin et al.[22]

*Signal-noise filtering.* Data were filtered to include RNAs with a min value > 0.5, max value < 20 across 5 RNA degradation conditions, and RNAs with a signal/noise ratio for SHAPE reactivity greater than 1.0. Signal/noise ratio for each construct was calculated as

$$SN\ ratio = \frac{1}{M}\sum_{i=1}^{M}\frac{1}{N}\sum_{j=1}^{N}\frac{\mu_{i,j}}{\sigma_{i,j}}, \qquad (4)$$

Where $\mu_{i,j}$ is the mean value of data type i at nucleotide j, and $\sigma_{i,j}$ is standard deviation of data type i at nucleotide j, as calculated by MAPseeker. The data that did not pass the above filters was also provided to participants to give the option to use in training, and was flagged with the variable `SN_filter=0`. Applying the above filter did not significantly alter the distribution of the median reactivity or signal/noise of any data type (SHAPE reactivity, deg_Mg_pH10, deg_Mg_50C) within either dataset (RYOS 1 or RYOS 2) (Figure S5). However, average signal/noise of the Round II constructs was higher than the Round 1 constructs. Average signal/noise for SHAPE reactivity increased from 5.3(2.4) to 6.2(3.5); for deg_Mg_pH10, 4.1(2.0) to 6.4(3.8) for Rounds 1 and 2; and for deg_Mg_50C 3.87(1.8) to 5.3(3.1) (Figure S2).

We wished to ascertain if the measured reactivities and degradation from Round 2 needed to be rescaled to match Round 1. To assess this, we compared distributions of nucleotide reactivities from nucleotide types. We found that for each data type and nucleotide type, the median values for Round 2 were within the 50% interquartile range of Round 1 (Figure S6A). We also compared distributions of reactivity from the first 5 nucleotides, which are a constant "GGAAA" for each construct. The median values for each nucleotide in this were within the 50% interquartile range for all except for the first two GG's in the `deg_Mg_pH10` data type (Figure S6B). We elected to not rescale the data from Round 2.

*Private test set curation.* The private test set was curated to avoid bias toward more highly-represented sequence motifs from the Eterna designs. Sequences that passed the above filters were clustered hierarchically using the `ward` method in scikit-learn[32] and then clustered at a cophenetic distance of 0.5. That is, sequences within the same cluster have <50% sequence similarity. All sequences that were in clusters with 1, 2, or 3 members were included in the private test set, as well as one cluster member from other randomly-selected clusters to attain the desired number of test set constructs.

*Comparing to the DegScore model.* We compared Kaggle models to the "DegScore" linear model[14], which models degradation at a given nucleotide *k* as a linear function of nucleotides surrounding *k*:

$$Y_k = \sum_{i=k-w}^{k+w} \sum_{n \in ACGU} \beta_{i,n} I_{i,n} + \sum_{i=k-w}^{k+w} \sum_{s \in HEIMBS} \beta_{i,s} I_{i,s} + \beta_0, \qquad (5)$$

Where *β* are the learned coefficients and I is an indicator function corresponding to the identity of nucleotide *k+i*. *I* accounts for sequence identity *n* (A,C,G,U) and its loop type assignment *s* (S=Stem, E=External Loop, I=Internal loop, B=Bulge, H=Hairpin, M=Multiloop, X=Dangle). *w* is the maximum window distance, set to be 12 in Leppek et al[14]. For a window size of w=12, or 25 positions, there are 251 parameters (25 positions with 4 sequence indicators and 6 secondary structure indicators for each position, and 1 intercept parameter).

*Data availability.* All datasets are downloadable in raw RDAT format from rmdb.stanford.edu at the following accession numbers: SHAPE_RYOS_0620, RYOS1_NMD_0000, RYOS1_PH10_0000, RYOS1_MGPH_0000, RYOS1_50C_0000, RYOS1_MG50_0000, RYOS2_1M7_0000, RYOS2_MGPH_0000, RYOS2_MG50_0000. Kaggle-formatted train and test sets are downloadable from https://www.kaggle.com/c/stanford-covid-vaccine. Datasets, scripts, and models are also included at https://www.github.com/eternagame/KaggleOpenVaccine.

*Code availability.* Code to run the Nullrecurrent model and the DegScore-XGBoost model is available at www.github.com/eternagame/KaggleOpenVaccine. Code to use and reproduce the linear regression DegScore model is available at www.github.com/eternagame/DegScore.

**Figures and Figure captions**

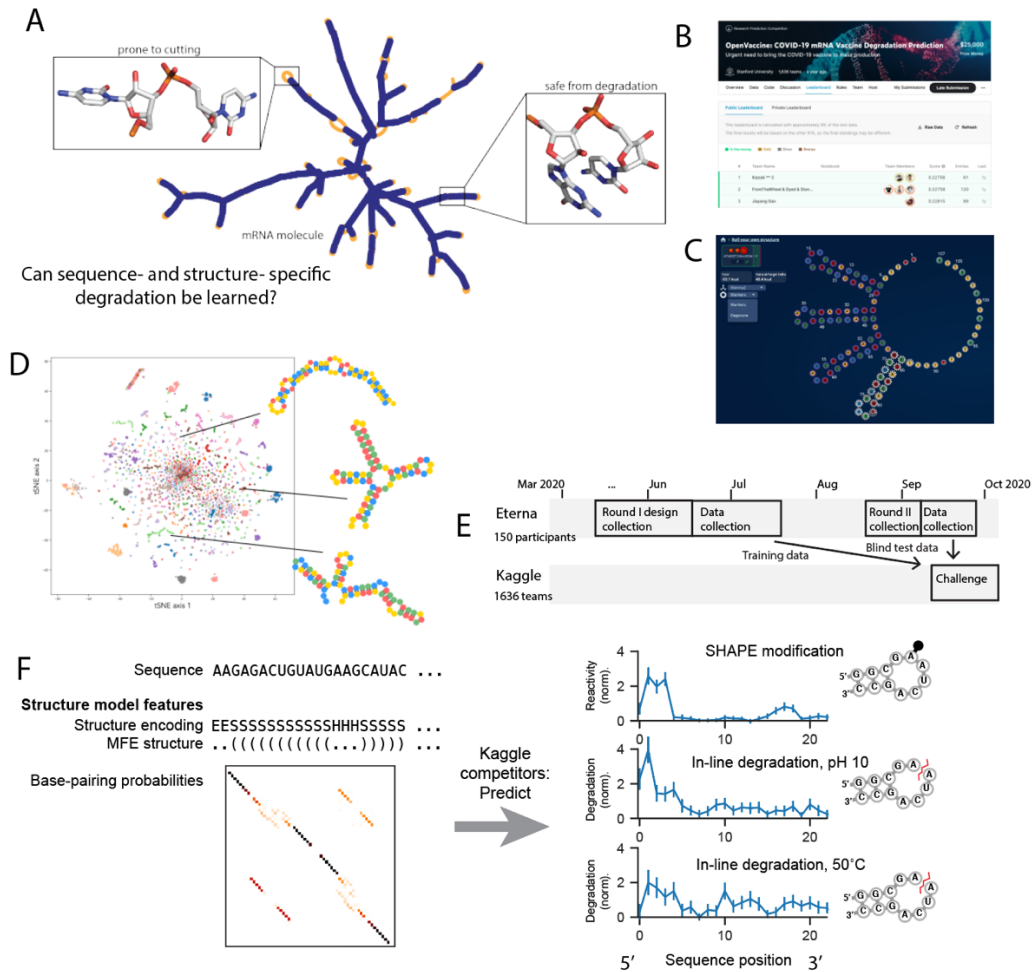

**Figure 1.** Dual-crowdsourcing setup for creating predictive models of RNA degradation. A. mRNA molecules fold into secondary structures containing unpaired regions prone to hydrolysis and limiting to therapeutic stability. B. Screenshot of the OpenVaccine Kaggle competition public leaderboard. C. Screenshot of an example construct designed by an Eterna participant in the "Roll Your Own Structure" challenge ("rainbow tetraloops 7" by Omei). D. tSNE[33] projection of training sequences of "Roll-Your-Own-Structure" Round I, marker style and colors indicating 150 Eterna participants. Lines indicate example short 68 nt RNA fragments. E. Timelines of dual crowdsourced challenges. Eterna participants designed datasets that were used for training and blind test data for Kaggle machine learning competition to predict RNA chemical mapping signal and degradation. F. Kaggle participants were given RNA sequence and structure information and asked to predict RNA degradation profiles and SHAPE reactivity. Error bars represent standard deviation.

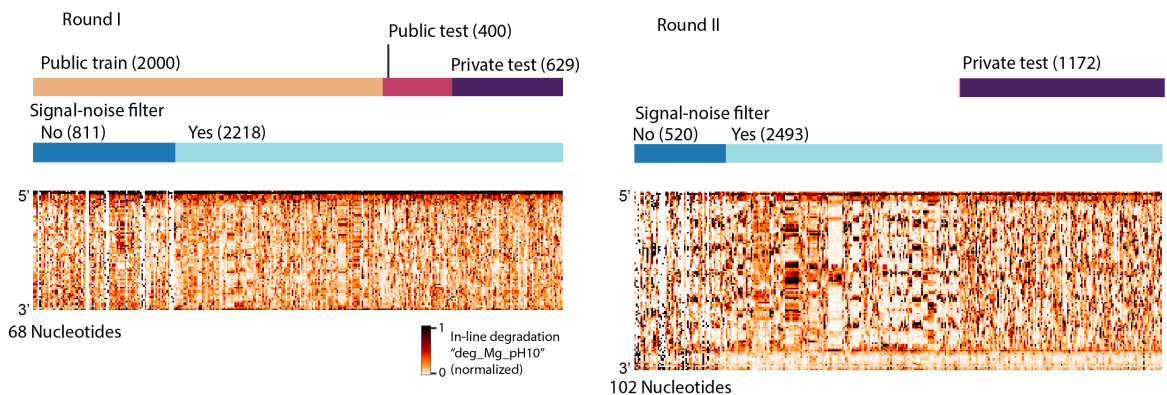

**Figure 2.** Signal-noise filtering and hierarchical clustering was used to filter the constructs designed by Eterna participants to create a test set of constructs that were maximally distant from other test constructs. Heatmaps of datatype "deg_Mg_pH10".

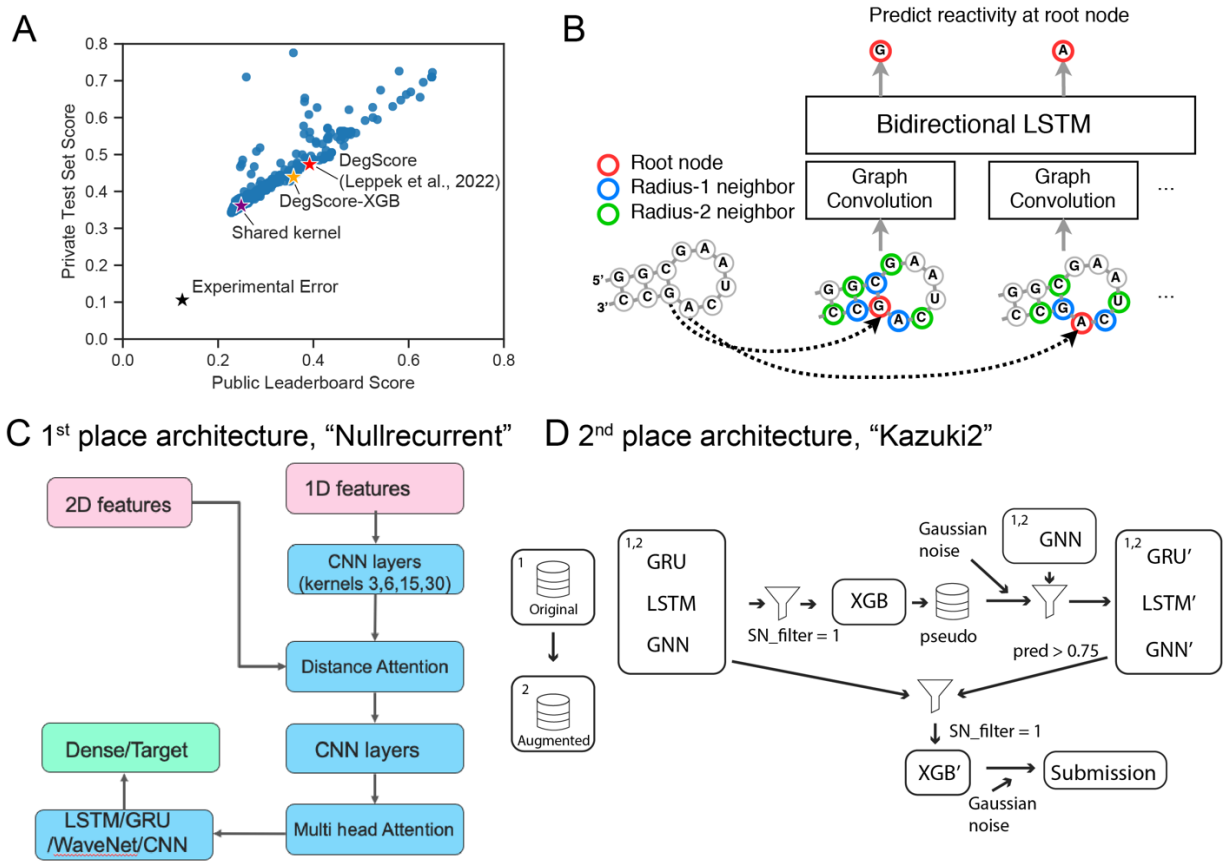

**Figure 3.** Deep learning strategies used in competition. (A) Public test vs. private test performance of all teams in Kaggle challenge. Black star: experimental error. Red star: DegScore baseline model[14]. Orange star: DegScore-XGB model using DegScore featurization with XGBoost. Purple star: baseline kernel used by many top-performing teams. (B) Distance embedding used to represent nucleotide proximity to other nucleotides in secondary structure. (C) Schematic of the single neural net (NN) architecture used by the first place solution. This solution combined two sets of features into a single NN architecture, which combined elements of classic RNNs and CNNs. (D) Schematic of the full solution pipeline for the second place solution. This solution combined single model neural networks, similar to the ones used for the first place solution, with more complex 2nd and 3rd level stacking using XGBoost[27] as the higher level learner.

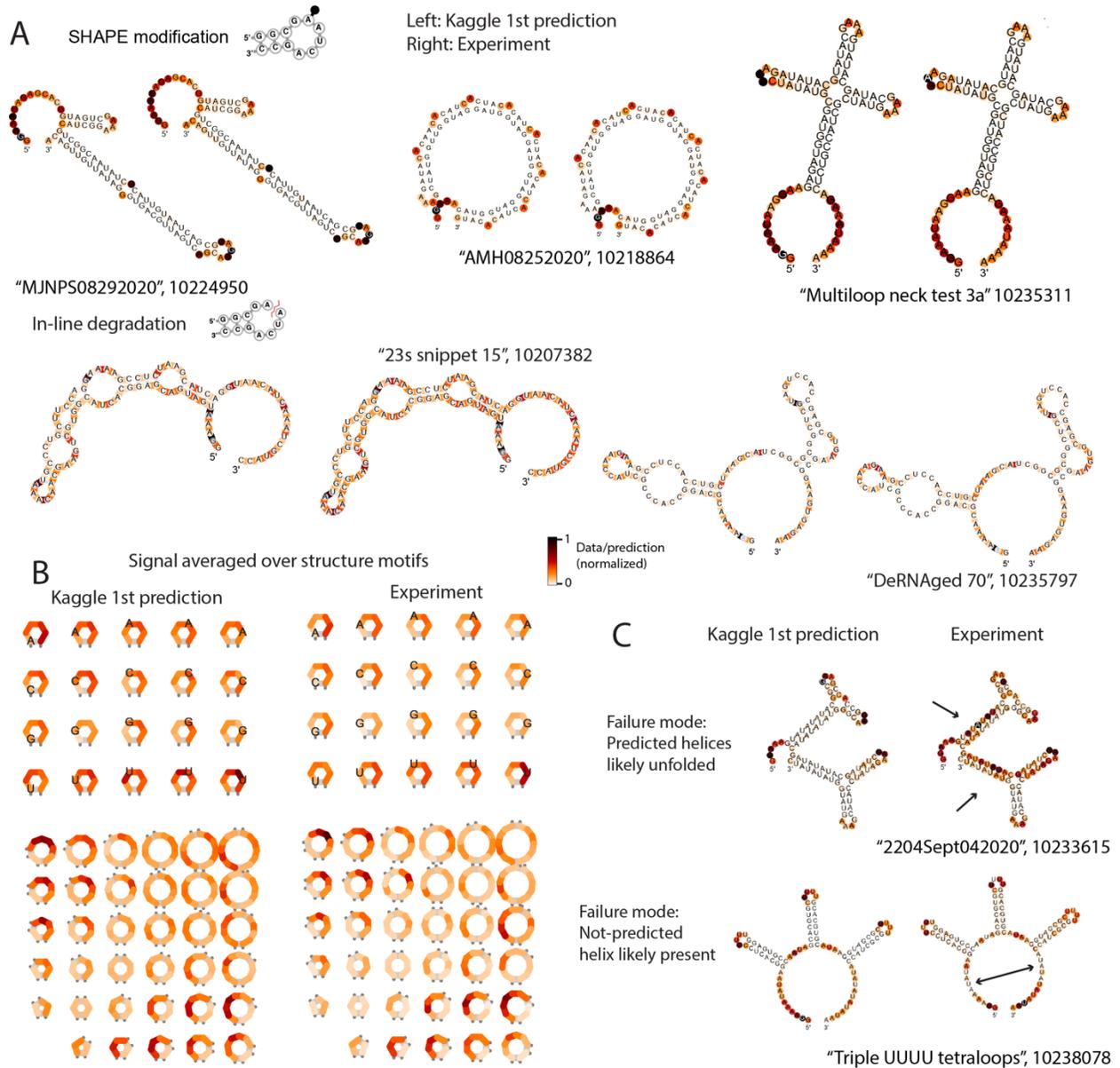

**Figure 4.** Deep-learning models can represent RNA-structure-based observables. (A) Representative structures from the best-predicted constructs from SHAPE modification (top row) and degradation at 10 mM $Mg^{2+}$, pH 10, 1 day, 24 °C (Deg_Mg_pH10, bottom row). (B) Nullrecurrent model predictions and experimental signal, averaged over secondary structure motifs. (C) One failure mode for prediction came from constructs whose input secondary structure features were incorrectly predicted.

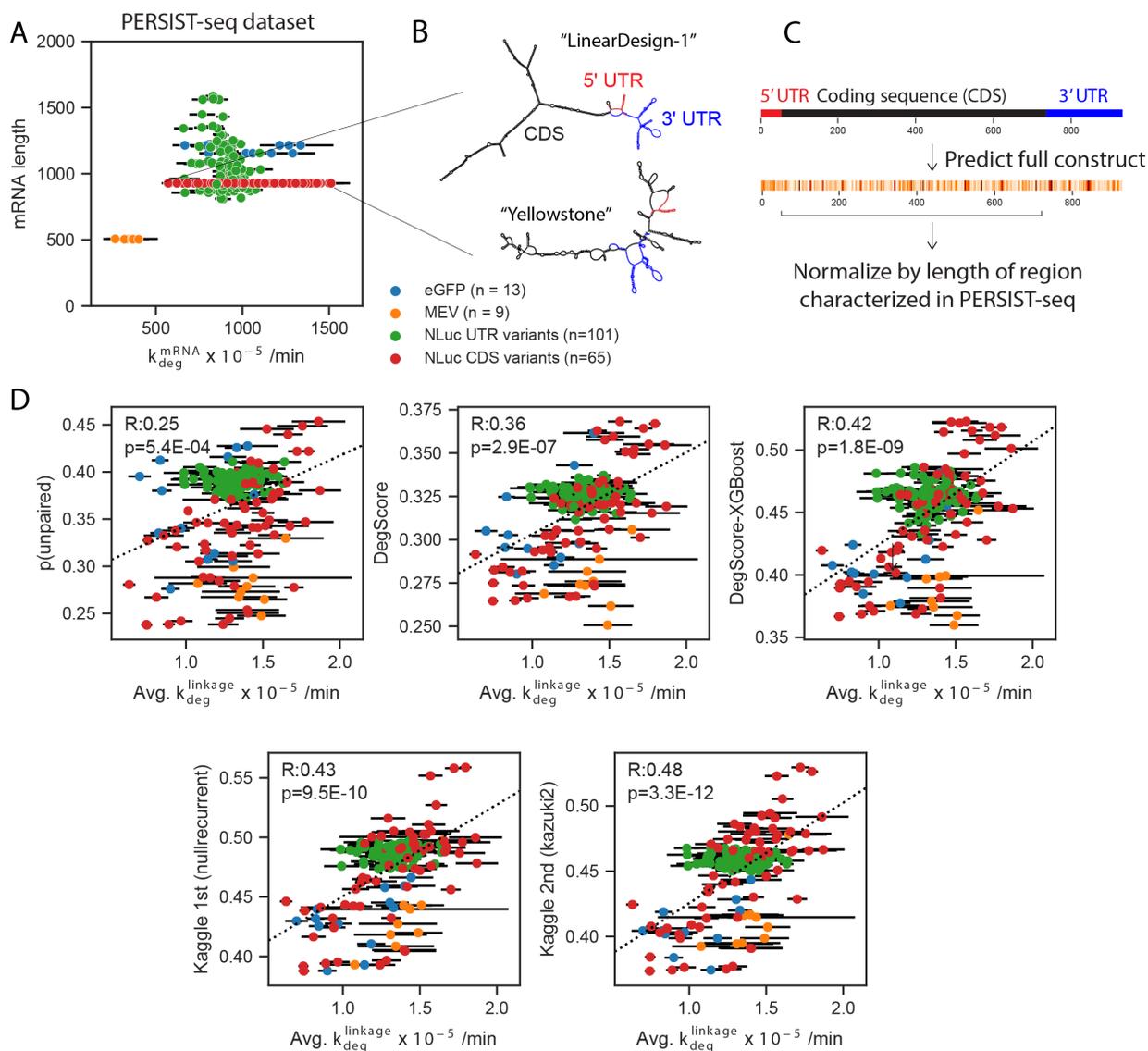

**Figure 5.** Kaggle models demonstrate improved performance in independent test of degradation of full-length mRNAs. (A) Overall mRNA degradation rate from PERSIST seq is driven by mRNA length. Kaggle models were therefore tested in their ability to predict length-averaged mRNA degradation. (B) Representative structures of two mRNAs of the same length that both encode Nanoluciferase, one with high degradation ("Yellowstone", left) and low degradation ("LinearDesign-1", right). (C) Prediction vectors were summed over nucleotides corresponding to the CDS region to compare to PERSIST-seq degradation rates, which account for degradation between two RT-PCR primers designed to capture degradation in the CDS region. (D) Length-normalized predictions from the Kaggle 1st place "Nullrecurrent" model and Kaggle 2nd place "Kazuki2" model show improved prediction over unpaired probabilities from ViennaRNA RNAfold[25] and the DegScore linear regression model[14], and a version of the DegScore featurization with XGBoost[27] training. Error bars represent standard error estimated from the PERSIST-seq experiment.

|  | Public test set (400 constructs, 27200 nucleotides) | Private test set (1801 constructs, 162316 nucleotides) | mRNA degradation prediction from ref. 6 (188 constructs) |
| --- | --- | --- | --- |
| **Metric** | MCRMSE | MCRMSE | Spearman Correlation |
| Experimental error | 0.12491 | 0.10571 | 0.88 |
| **Single Model (blind prediction)** | | | |
| DegScore | 0.39219 | 0.47297 | 0.36 |
| DegScore-XGBoost | 0.35854 | 0.43850 | 0.42 |
| Nullrecurrent | 0.22758 | **0.34198** | 0.43 |
| Kazuki2 | **0.22756** | 0.34266 | **0.48** |
| **Ensembled models (post hoc)** | | | |
| Genetic algorithm (10 of top 100 selected) | **0.2237** | 0.3397 | |
| Ensemble top 2 models | 0.2244 | **0.33788** | 0.47 |
| Genetic algorithm on private test set | | 0.3382 | -- |

**Table 1.** Results from models tested in this work on Kaggle OpenVaccine public leaderboard, private test set, and orthogonal mRNA degradation results. [1]Bootstrapped Spearman correlation of degradation rate (resampled from experimental error) to half-life.

**Supporting information corresponding to "Diverse machine learning models for predicting RNA degradation via dual crowdsourcing"**

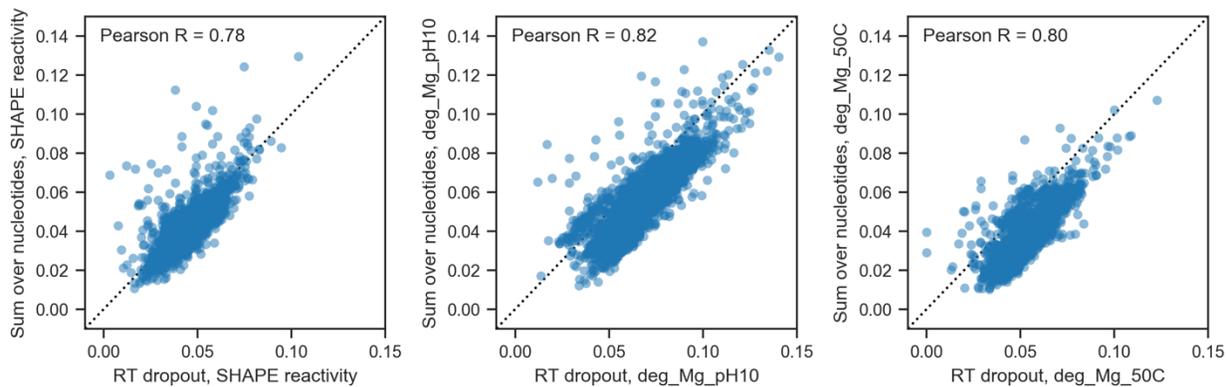

**Figure S1.** Summed per-nucleotide degradation rates and overall degradation rate, estimated by log(abundance of full-length construct), are highly correlated in Rounds 1 and 2.

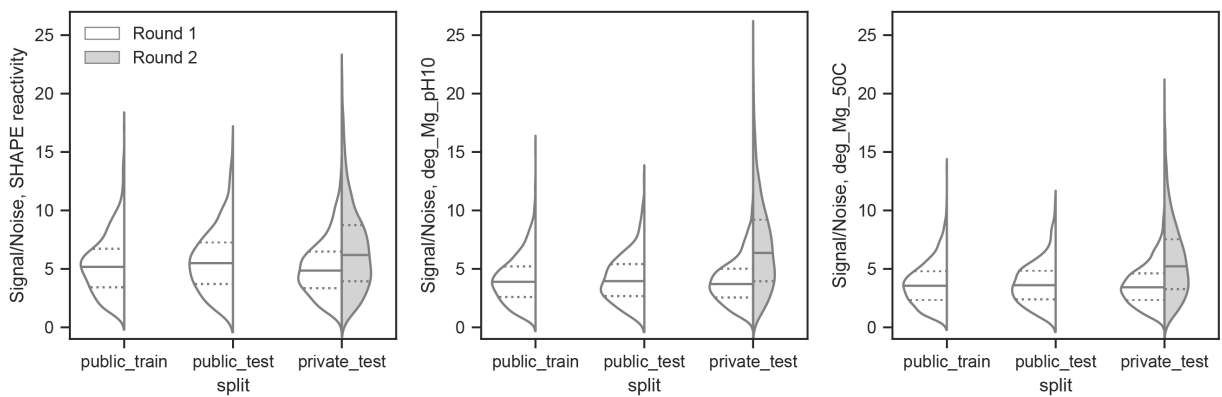

**Figure S2.** Average signal-noise from data splits, separated by data from Rounds 1 and 2. Solid lines: median, dotted lines: 25/75% percentile.

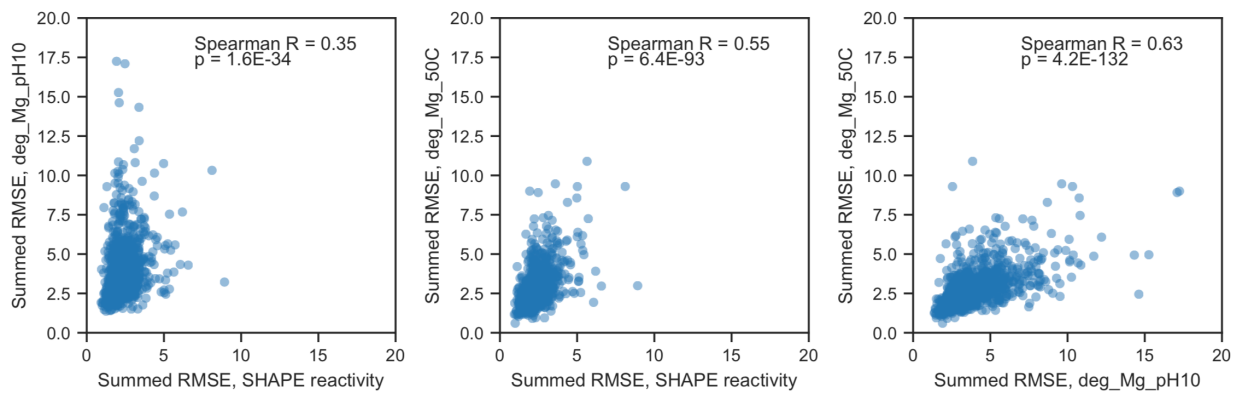

**Figure S3**. Correlation between RMSE by data type across constructs.

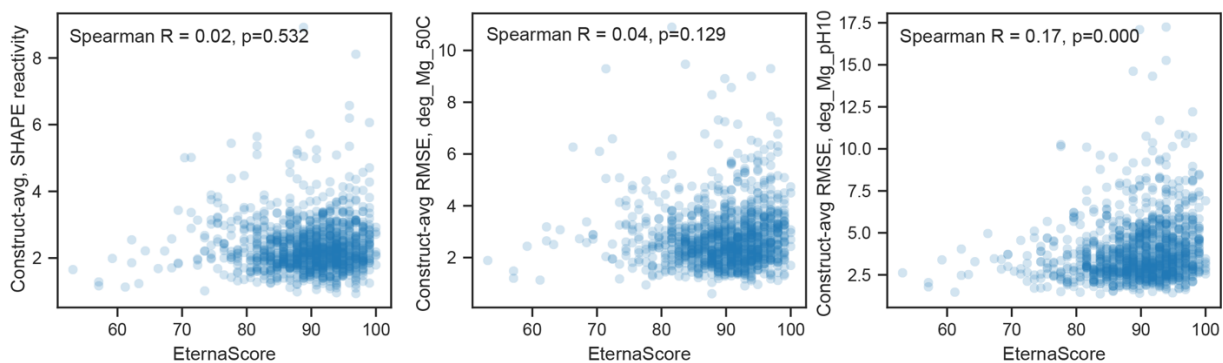

**Figure S4.** RMSE of Nullrecurrent model did not correlate with EternaScore, a measure of how closely the SHAPE reactivity data matched the predicted secondary structure.

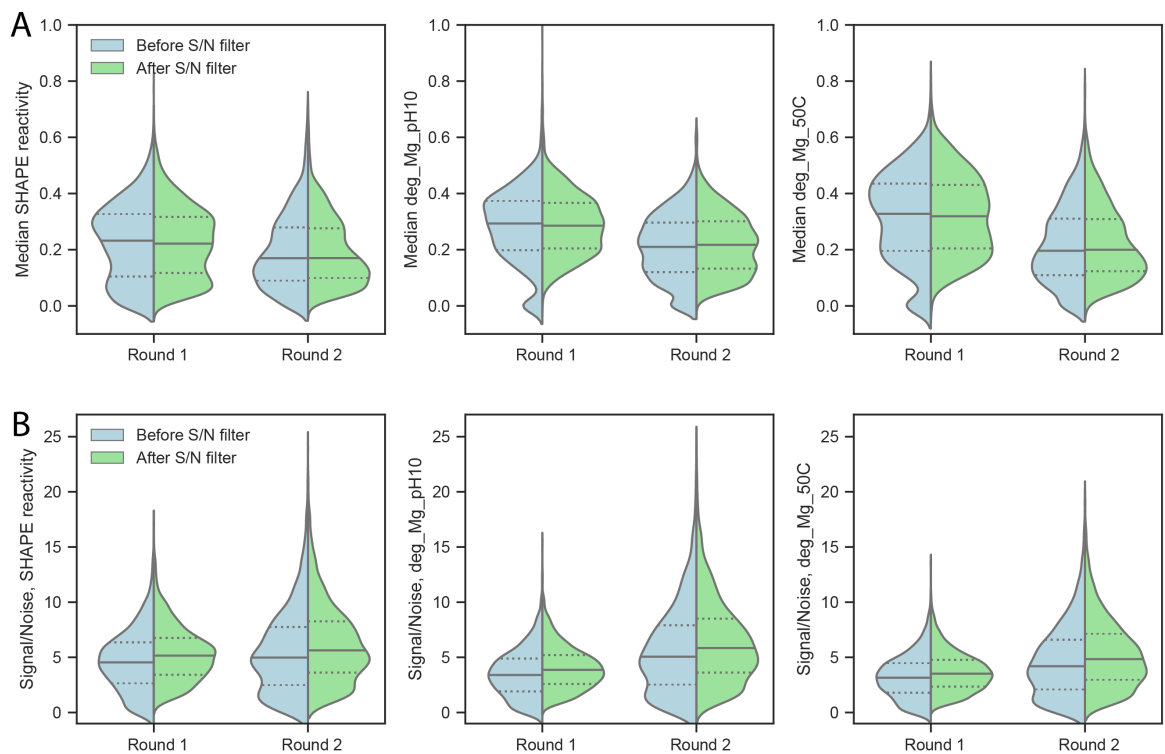

**Figure S5.** Effect of signal-noise filter (described in Methods) on (A) median reactivity/degradation per construct and (B) average signal-noise per construct for Rounds 1 and 2. Solid lines: median, dotted lines: 25/75% percentile.

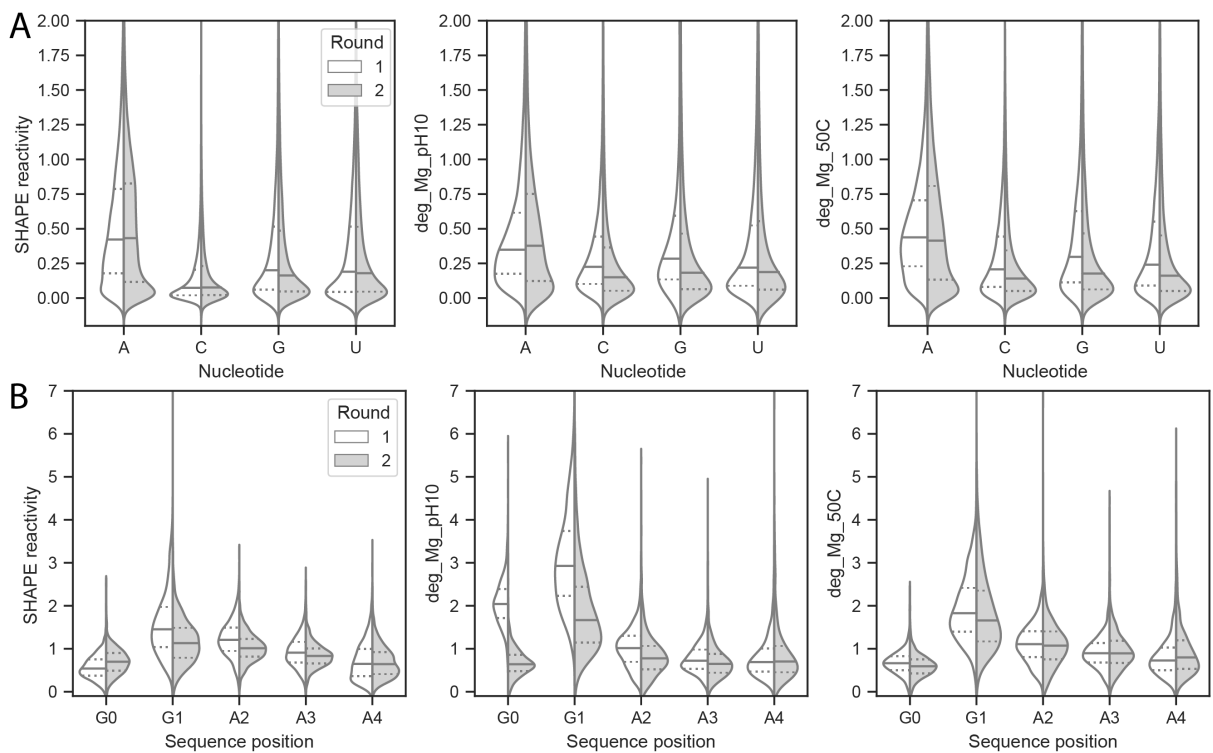

**Figure S6. (A)** Reactivity/degradation data disaggregated by nucleotide from rounds 1 and 2, from constructs that passed signal-noise filter. (B) Reactivity/degradation data from first 5 nucleotides from rounds 1 and 2, which were a constant "GGAAA". Solid lines: median, dotted lines: 25/75% percentile.

**Solution descriptions from Kaggle teams**

1st Place - Jiayang Gao (also called "Nullrecurrent" model)
My final submission is a simple ensemble of 4 models, each starts with an auto-encoder pretrained GNN architecture, followed by LSTM, GRU or Wavenet layers. My model takes in two kinds of features, (1) 1D features with length n (the length of the mRNA sequence), representing features at each location of the sequence; (2) 2D features with size n^2, representing "distance" or "relationship" concepts between each pair. The strongest 1D features are the distance to the closest paired position, as well as the distance to the closest unpaired position - reactivity increases as the distance to the closest paired position becomes larger. The strongest 2D feature is the distance between each pair in the primary pairing based graph - this allows the model to capture the "neighbors" caused by pairing. One major difficulty of this problem is generalizing the model to different sequence lengths, and I use a semi-supervised approach to make my model more generalizable. In particular, I randomly generate sequences of different length, calculate their BPP matrix and pairing structure using the Arnie library, pseudo label their targets, and train them together with the original train dataset. This semi-supervised learning approach is particularly useful in this problem, for which sample size is small and labeled data is expensive to obtain.
https://www.kaggle.com/c/stanford-covid-vaccine/discussion/189620

2nd Place - Kazuki ** 2 (also called "Kazuki2" model)
The mRNA sequence is not just a linear series of data, but also constitutes a loop by pairing between specific bases. Therefore, we thought of constructing LSTM/GRU and GNN independently, and integrating the prediction results with XGBoost. The base pair probabilities (bpps) are calculated by prediction, and the calculation results differ depending on the algorithm used. Therefore, we used several algorithms in Arnie to predict base pairs, and used these algorithms as data augmentation and input multiple bpps simultaneously to improve the performance. We prepared 38 LSTM/GRU-based and 49 GNN-based bpps by changing the types of bpps and their architectures, and integrated them with XGBoost to help improve the stability of prediction. Also, since the sequences in the test set tended to be longer than those in the training data, we confirmed in preliminary experiments that the model trained by shortening the sequences in the training data (107 to 88) was applicable to sequences of the original length (107).
https://www.kaggle.com/c/stanford-covid-vaccine/discussion/189709

3rd Place - Striderl
My solution is an ensemble of various models of different structures and different training techniques. I added various LSTM/GRU/wavenet layers at the end of the AE pretrained GNN structure, in which 2 x 128 units of LSTM or GRU layers work the best for me. Besides the common features used by other teams like structure adjacency matrix and neighbor adjacency matrix, I used RNAComposer to generate 3D structures for each sample in the competition, and used predicted 3D distance to form the distance matrix. I used Arnie to predict other possible base pairs as data augmentation. Besides, another augmentation I used was to reverse the sequence and targets. The two augmentations quadruple the size of the original data. I also used pseudo labeling technique to iteratively improve my best single model.

4th Place - FromTheWheel & Dyed & StoneShop
The 4th place solution is a blend of 4 different models, in which the RNN layers were varied (LSTM+LSTM, LSTM+GRU, GRU+LSTM and GRU+GRU). We represent the RNA sequences as graphs where each base corresponds to a node. The network then learns a representation for each of these graphs and passes this representation through bi-directional RNN layers to obtain a sequence of predicted targets. Both the edge and node features were derived from the given sequence and provided Base Pairing Probability (BPP) matrix. One-hot-encoded bases (A, G, C, or U), one-hot-encoded positional feature (the remainder of the base index divided by 3), one-hot-encoded loop types, loop type probabilities (CapR) and BPP sum and number of zero's were beneficial node features. The distances (manhattan) between bases, normalized by sequence length and whether there is a base-pairing indicated in the structure were used as edge features. All these features were also inferred for newly generated BPP matrices, generated with 6 libraries available within the ARNIE software package. The new information derived from the CONTRAfold library was the most useful for this task, followed by RNAsoft, RNAstructure and Vienna. We also tried to use 3D angle information (binned and then categorically encoded) extracted with AMIGOS: this boosted our simple LSTM architecture used to fast check the feature importance but deteriorated the performance of our final model.

5th Place - tito
My solution is a simple ensemble of GNN-based model and GRU/LSTM-based model. The features are not significantly different from those used by the other teams. I focused mainly on augmentation. (1) I used eternafold, vienna, nupack, contrafold and rnasoft to extract structure and loop_type. These backend engines are used to extract additional bpps too. Especially eternafold and contrafold worked well. (2) In this competition, the sequence length of the test data was longer than that of the training data, so we added a dummy sequence to the training data. (3) I added reversed sequence to training data. I think augmentation was important in this competition.

6th Place - nyanp
The data in this competition is very unique in two ways: 1) the sequence length is different between the training data and the private test data, and 2) the sequence has long-term dependencies via pairing. My NN model is constructed by stacking 1D SE-ResNet Layer and Graph Convolution Layer in order to make the model invariant to sequence length and to capture the long-term dependency. The Graph Convolution Layer is computed by a simple sum of products of the BPP matrix or adjacency matrix and the sequence feature vector. The best single model was ranked 42nd (0.35045) on the private leaderboard, while it was 513th (0.24371) on the public leaderboard. This indicates that my model is more robust to changes in sequence length compared to the other participants. Best ensemble achieved MSRMSE's of 0.23069/0.34538 on the public/private leaderboard by combining 4 different architectures.
https://www.kaggle.com/c/stanford-covid-vaccine/discussion/189241
https://www.kaggle.com/nyanpn/6th-place-cnn-gcn

7th Place - One architecture
Main features for my solution used were base pairing probability matrix, nucleotide sequence, structure, and loop type. Additionally, an inverse distance matrix (nucleotides at position i and j have distance |i-j| between them) was added to the base pairing probability matrix before inputting it as a bias for the self-attention matrix. The conventional type of positional encoding as detailed in the original transformer or learnable positional encoding were not used; instead, position was encoded by the inverse distance matrix. Further, 5 secondary structure packages were used to generate pairing probability matrix, structure, and loop type at 37 and 50 C, resulting in 10 sets of features for each sequence. The architecture used was almost identical to bert aside from the 1D/2D convolution/deconvolution layers (without padding). The core module (ConvTransformerEncoder) was constructed as follows: (1) 1D convolution on the sequence of encodings and 2D convolution on the bpp feature map. (2) self-attention with bpp feature map as additive bias (3) position wise feedforward network. (4) 1D deconvolution on the sequence of encodings and 2D deconvolution on the bpp feature map. All available sequences were used to pretrain (unsupervised) models on randomly mutated or masked (with NULL token) sequence retrieval loss (basically just softmax to retrieve correct nucleotide/structure/loop). For convenience, two linear decoders were initialized before pretraining, one for sequence retrieval, and another for degradation predictions later on. The Ranger optimizer was used with a flat and anneal schedule. Some sequences were excluded during training on degradation targets based on signal to noise threshold (0.25, 0.5, or 1). My biggest discovery from this competition is that the vanilla positional encoding used in the original transformer paper does not generalize well to this task at least. It seems that the type of positional encoding used in most transformers does not adequately describe the concept of position, which is fine for NLP because I believe order and position of words are not as important as for RNA. The vanilla positional encoding is more of an absolute positional encoding, whereas the inverse distance basically encoders relative position in a very simple way that generalized better to longer sequences. Best ensemble achieved MSRMSE's of 0.23056/0.34550 on the public/private leaderboard.

8th Place - ishikei
My solution is an ensemble of GRU/LSTM and GNN. Each model is AE pretrained with all data. For features, bpps was augmented with different temperature parameters (T=37, 50) using the ARNIE package (vienna, nupack, rnastructure, rnasoft, eternafold, contrafold). I also added the shannon entropy at each base position. Because all of the data in this competition are predicted values except for the sequence, I think it would have been effective to use an ensemble of bpps predictions from various algorithms. At high temperature: Since the secondary structure of RNA is temperature-dependent, I think it was effective to use bpps with T changed as input. At high pH: Alkaline hydrolysis of RNA can occur at any position in the sequence (probably) as well, so I think the prediction itself is difficult.
https://www.kaggle.com/c/stanford-covid-vaccine/discussion/190314

9th Place - Keep going to be GM
The big issue of this contest is that the train and test RNA sequence lengths are different, and the data contains noise. The RNA sequence length used for training was 107, and the final ranking was obtained for 130 sequences. Different models with fluent feature engineering to enhance the generalization of predictions. Because RNA can have both graph and sequence, traditional recurrent neural networks (LSTM, GRU), transformer and graph-neural networks were applied. For feature engineering, (A, G, C, or U) were represented by embedding layers. And, various (N, N) adjacency matrices called bpps, which are probabilities of being linked between nucleotides are calculated with various softwares, such as CONTRAFold, RNAFold. The statistical features such as ratio of (A, G, C, or U) in sequence are added. has are also included.I created the (N, N) matrix for fixed attention in various ways. Bpps was created using several packages. The distance matrix was augmented using gaussian. With these features, dozens of models were created by differing hyperparameters of sequential blocks and graph convolution layers.
https://www.kaggle.com/c/stanford-covid-vaccine/discussion/189845

11th Place - Social Distancing Please
Our solution is an ensemble of multiple models. There are mainly 2 types of models. The first type is a combination of 1D Convolution Layers, Graph Convolution Layers, and RNN Layers. The second type is a combination of WaveNet layers and RNN Layers. The most powerful features are the adjacency matrices constructed by the given structure sequence and the given base-pair probabilities. The adjacency matrices are used for the Graph Convolution Layers. Another useful trick we have used is to apply a lower training weight to the top 6 sequence position, it is because the sequence start is similar across the different sequences. We also used 2 different Linear Regression models to ensemble the predictions in different sequence positions that are seqpos[:6] and seqpos[6:].

13th Place - The Machine
The main idea behind our approach is generating bpp matrices from all the libraries included in Arnie, training a model from the output structures of each library and finally creating an ensemble of all the trained models. Although each library provides sub optimal bpp, their consensus provides a better solution. We also included several architectures in the ensemble and the best of them consisted of 1D convolution, static graph convolution and bi-directional LSTM layers. A static graph convolution layer processes each two connected nucleotides in the predicted secondary structure (zeros added when a nucleotide isn't connected). All our models were trained on all the data using self supervised learning then fine-tuned on the training data only using supervised learning.